# A Logic for Default Reasoning About Probabilities

Manfred Jaeger
Max-Planck-Institut für Informatik,
Im Stadtwald, 66123 Saarbrücken

## Abstract

A logic is defined that allows to express information about statistical probabilities and about degrees of belief in specific propositions. By interpreting the two types of probabilities in one common probability space, the semantics given are well suited to model the influence of statistical information on the formation of subjective beliefs. Cross entropy minimization is a key element in these semantics, the use of which is justified by showing that the resulting logic exhibits some very reasonable properties.

## 1 INTRODUCTION

It has often been noted that "probability" is a term with dual use: it can be applied to the frequency of occurrence of a specific property in a large sample of objects, and to the degree of belief granted to a proposition.

While some have argued that only one of these two interpretations captures the true meaning of probability [Jay78], others have tried to analyze both usages of the term in their own right, and to clarify the relationship between the two aspects of probability.

Carnap was among the first to do this ([Car50]). Even though his interest lies primarily with probabilities as subjective beliefs (or "degrees of confirmation"), he also formulates *direct (inductive) inference* as a principle to arrive at subjective beliefs on the basis of given relative frequencies: when it is known that objects from a class $C_1$ also are members of a class $C_2$ with a frequency $p$, and a specific object $a$ is believed to belong to $C_1$, then the given statistical information may be used as a justification for assigning $p$ as a degree of belief to $a$'s belonging to $C_2$.

When, instead of firmly believing that $a$ is an element of $C_1$, one only has several conflicting pieces of evidence about the true nature of $a$, these can be combined to form a degree of belief for $a$ being in $C_2$ by using *Jeffrey's rule* [Jef65], as illustrated in the following example.

**Example 1.1** Scanning channels on TV we have tuned in to a mystery film. It looks interesting, but we only want to continue watching it, if a happy ending seems likely.

By what we have seen so far, we judge the film to be either American, French, or English, with a likelihood of 0.2, 0.6, and 0.2 respectively. From our extensive past experience with mystery films we know that 8 out of 10 American films have a happy end, while this figure is 1 and 5 out of 10 for French and English productions, respectively.

Jeffrey's rule in this situation states, that our degree of belief in a happy ending of the film we are currently watching should be given by

$$0.2 \times 0.8 + 0.6 \times 0.1 + 0.2 \times 0.5 = 0.32. \quad (1)$$

Hence, we better switch channels.

It must be noted at this place that calling the inference in this example by the name of Jeffrey's rule is a somewhat loose terminology: when Jeffrey originally stated his rule he was concerned with updating prior subjective beliefs to posterior subjective beliefs in the light of newly obtained evidence, not with using statistical information to define degrees of belief. Hence, the terms for the conditional probabilities of a happy end given the origin of a film that appear in (1) would be some prior conditional beliefs in Jeffrey's rule, rather than statistical expressions. If, however, the fundamental assumption is made that in the absence of any specific information about an object, the subjective beliefs held about the object are governed by the statistical information available for the domain from which it is taken, then conditional belief and conditional statistical probability can be equated, and the rule for deriving degrees of belief from statistical information exemplified by (1) be identified with Jeffrey's rule.

Note, too, that this assumption also underlies the direct inference principle, which from this perspective then can be seen as a special case of Jeffrey's rule — special both in the way prior beliefs are defined through relative frequencies and in that the new evidence concerning the nature of an object takes the form of one certain fact.

The kind of probabilistic inference illustrated in example 1.1 might be called *default reasoning about probabilities*. While this should be clearly distinguished from logical default reasoning (e.g. [McC80], [Rei80]), it shares the nonmonotinicity of these logics: in the light of additional (probabilistic or definite) information, earlier inferences may be retracted.

Recently, proposals have been made to incorporate the two kinds of probabilistic statements in an extension of first-order predicate logic [Hal90], [Bac90]. Here, statistical information and subjective beliefs are modelled by probability measures on the domains of first-order structures and sets of possible worlds, respectively. While this is an intuitively appealing interpretation of the formulas, it does not allow for the kind of reasoning exemplified by Jeffrey's rule. The probability measures on the domain and the possible worlds can be chosen independently in such a way that all the formulas in the given knowledge base are satisfied, but no interaction between the two kinds of probabilistic statements takes place. Indeed, Halpern writes [Hal90]: "Although $\mathcal{L}_3(\Phi)$ [the combined probabilistic logic in question] allows arbitrary alternation of the two types of probability, the semantics does support the intuition that these really are two fundamentally *different* types of probability."

An additional strategy to arrive at subjective beliefs on the basis of statistical information is developed in [Bac91], [GHK92a], [GHK92b], [BGHK92], and [BGHK93]. This strategy, which is based on direct inference, has the great disadvantage that it does not allow for any given subjective beliefs to be used for arriving at new degrees of belief. Hence, even Jeffrey's rule is beyond the scope of reasoning that can be carried out in this framework. On the other hand, very specific degrees of belief are assigned to propositions even in the absence of any information: on the basis of an empty knowledge base, the propositions American($this\_film$) and American($this\_film$)∧ Happy_end($this\_film$) would be assigned a degree of belief of 0.5 and 0.25 respectively.

The formalism presented in this paper, though motivated by similar intuitions as the above mentioned, exhibits rather different properties. Among them are:

- The expressive power of the language used is smaller than in [Hal90], [Bac90]. Notably, expressions about statistical and subjective probabilities can only be combined in a restricted way.

- Both statistical and subjective probabilities can be specified in a knowledge base, and new probabilities of both types be inferred. While the statistical probabilities entailed by the knowledge base essentially depend on the given statistical information only, the resulting subjective beliefs depend crucially on both types of probabilistic statements.

- When only partial statistical information is available (as is usually the case), no default assumptions about the statistical probabilities are made. As a result, it will usually only be possible to infer probability intervals rather than unique probability values from a knowledge base

The basic idea on which the formalism to be defined in the following sections is based, is to interpret both types of probabilistic expressions by probability measures on a common probability space. In the example above it is noticeable that both the statements about the relative frequency for happy endings and the subjective assignment of likelihood to the predicates American, French and English, are basically constraints on a probability measure on the formulas in the vocabulary S= {American, French, English, Happy_end}, where in some cases a constant "$this\_film$" enters as a parameter. When deductions from the knowledge base are made, it is again probabilities on these formulas that are to be inferred. Default reasoning about probabilities can now be viewed as the process of selecting a probability measure on the expressions $this\_film \in \phi$ (with $\phi \in L_S$, i.e. a first-order formula over S) that most closely resembles the probability measure generally assigned to $L_S$ on the basis of the given statistical information.

While it would be desirable, to work with probability measures on the abstract syntactic structure $L_S$ itself (as has been done for terminological logics in [Jae94]), it proves much easier, in the more general framework of first-order predicate logic, to use probability measures on the domain of an interpretation to induce a probability measure on $L_S$.

## 2 SYNTAX

As mentioned above, it will not be possible to freely combine expressions about the two different kinds of probabilities. Hence, two distinct extensions of the syntax of first-order logic have to be provided.

The following notational conventions will be used in the sequel: tuples $(v_0, \ldots, v_{n-1})$, $(a_0, \ldots, a_{k-1})$ of variable or constant symbols are abbreviated by $v$, $a$. When it is necessary to explicitly note the length of a tuple, the notation $\overset{n}{v}$, $\overset{k}{a}$ may be used. $\phi(v)$ is used to denote a formula $\phi$ whose free variables are among $v_0, \ldots, v_{n-1}$.

**Definition 2.1** Let S be a vocabulary containing relation-, function-, and constant-symbols. A *statistical formula in* S is any formula that can be constructed from S by the syntax rules of first-order predicate logic with equality together with the new rule:

- If $\phi(v)$ and $\psi(v)$ are statistical formulas in S, $\{v_{i_1},\ldots,v_{i_k}\} \subseteq \{v_0,\ldots,v_{n-1}\}$, and $p \in [0,1]$, then

$$[\phi(v) \mid \psi(v)]_{(v_i)}^k \geq p$$

is a statistical formula in S. The free variables in this formula are the free variables of $\phi$ and $\psi$ without $\{v_{i_1},\ldots,v_{i_k}\}$.

The set of statistical formulas in S is denoted by $L_S^\sigma$. A statistical formula with no free variables is a *statistical sentence*.

**Definition 2.2** Let S be as above, and $\{a_0,\ldots,a_{n-1}\}$ a set of constant symbols not in S. A *subjective probability sentence for $a$ in* S is any sentence of the form

$$\text{prob}(\phi[a] \mid \psi[a]) \geq p$$

with $\phi(v), \psi(v) \in L_S^\sigma$, and $p \in [0,1]$. $L_S^{\beta(a)}$ denotes the set of all these sentences.

The abbreviations $[\phi \mid \psi]_{(...)} \leq p$ and $\text{prob}(\phi \mid \psi) \leq p$ may be used for $[\neg\phi \mid \psi]_{(...)} \geq 1-p$ and $\text{prob}(\neg\phi \mid \psi) \geq 1-p$ respectively.

Similarly, $[\phi \mid \psi]_{(...)} = p$, $[\phi \mid \psi]_{(...)} < p$ and $[\phi \mid \psi]_{(...)} > p$ are defined. Also, $\text{prob}(\phi \mid \psi) = p$ may be substituted for the pair of sentences $\text{prob}(\phi \mid \psi) \geq p$ and $\text{prob}(\phi \mid \psi) \leq p$. Note, however, that $\text{prob}(\phi \mid \psi) < p$ would have to be defined by means of the negation of $\text{prob}(\phi \mid \psi) \geq p$, and such a negation is not within the syntax given by definition 2.2. Finally, $[\phi]_{(...)} \geq p$ and $\text{prob}(\phi) \geq p$ are used for $[\phi \mid \tau]_{(...)} \geq p$ and $\text{prob}(\phi \mid \tau) \geq p$, where $\tau$ is any tautology.

Definition 2.1 is standard and can be found similarly in [Kei85], [Hal90], [Bac90]. Definition 2.2 differs from its counterparts in [Hal90] and [Bac90] in that $\text{prob}(\phi \mid \psi) \geq p$ is seen as a statement about a distinguished subset of the constant symbols appearing in $\phi$ and $\psi$, and $\phi, \psi$ are not allowed to contain, in turn, a formula of the form $\text{prob}(\phi' \mid \psi') \geq p'$.

A knowledge base KB in the language here defined consists of a finite set $\Phi^\sigma$ of statistical sentences (which will typically also contain some purely first-order sentences), and a finite set $\Phi^{\beta(a)}$ of subjective probability sentences for $a$.

**Example 2.3** The probabilistic knowledge from our introductory example can be symbolized by a knowledge base $KB_{f1} = \Phi^\sigma_{\text{Movies}} \cup \Phi^{\beta(f1)}$ where $f1$ is a constant symbol standing for the unknown film we are concerned with, and

$\Phi^\sigma_{\text{Movies}} =$

[Happy_end$v$ | American$v \wedge$ Mystery$v]_{(v)} = 0.8 \quad (2)$

[Happy_end$v$ | English$v \wedge$ Mystery$v]_{(v)} = 0.5 \quad (3)$

[Happy_end$v$ | French$v \wedge$ Mystery$v]_{(v)} = 0.1 \quad (4)$

$\Phi^{\beta(f1)} :=$

prob(American$f1 \wedge$ Mystery$f1) = 0.2 \quad (5)$

prob(English$f1 \wedge$ Mystery$f1) = 0.2 \quad (6)$

prob(French$f1 \wedge$ Mystery$f1) = 0.6 \quad (7)$

## 3   SEMANTICS

### 3.1   OUTLINE

A semantical structure in which $\Phi^\sigma$ can be interpreted is defined along the same lines as in [Kei85], [Hal90], [Bac90]:

**Definition 3.1** A *statistical S-structure* is a structure $(\mathcal{M}, I)$, where

- $\mathcal{M} = (M, \mathfrak{M}, \mu)$ is a probability space with domain M, supplied with a $\sigma$-algebra $\mathfrak{M}$ and a probability measure $\mu$ on $\mathfrak{M}$.

- I is an interpretation function that maps the relation-, function-, and constant-symbols in S to relations, functions and elements of M in such a way that for every formula $\phi(v) \in L_S^\sigma$, the interpretation $I(\phi) \subseteq M^n$ is measurable with respect to the product $\sigma$-algebra $\mathfrak{M}^n$.

The condition imposed on I in this definition may seem highly restrictive; in fact, one may wonder whether statistical structures in the sense of definition 3.1 actually exist. While it is beyond the scope of this paper to embark on a thorough measure-theoretic discussion of the questions here involved, it should be pointed out, that whenever M is finite or countably infinite, then $\mathfrak{M}$ can be taken to be the power set $2^M$, and every subset of $M^n$ is measurable with respect to $\mathfrak{M}^n$.

What about an interpretation for $\Phi^{\beta(a)}$? As was indicated in the introduction, subjective probability sentences can be seen as making assertions about a probability measure on the formulas in the vocabulary S, with $a$ just a name or parameter for this measure. Put another way, in the context of a fixed structure $(\mathcal{M}, I)$, $a$ in this interpretation may be viewed as a random variable with values in $M^n$, and $\Phi^{\beta(a)}$ is a set of constraints on its distribution. Since these constraints only concern subsets of $M^n$ that are definable by formulas in $L_S^\sigma$, which, by definition 3.1, all belong to $\mathfrak{M}^n$, this leads to the following definition.

**Definition 3.2** A *probabilistic S-structure for $a$* is a structure

$$(\mathcal{M}, I, \nu_a),$$

where $(\mathcal{M}, I)$ is a statistical S-structure, and $\nu_a$ is a probability measure on $\mathfrak{M}^n$.

Some conditions are immediate for when a probabilistic S-structure shall be called a *model* of a knowledge base $KB = \Phi^\sigma \cup \Phi^{\beta(a)}$: for a statistical S-structure

$(\mathcal{M}, I)$, a valuation function v, and a statistical formula $\theta$, the relation

$$((\mathcal{M}, I), v) \models \theta$$

is defined by augmenting the standard definition for first-order logic with the rule

$$((\mathcal{M}, I), v) \models [\phi(v) \mid \psi(v)]_{(v_i^k)}^k \geq p \quad \text{iff}$$

$$\mu^k(\{(m_1, \ldots, m_k) \in M^k \mid$$
$$((\mathcal{M}, I), v[\overset{k}{m} / \overset{k}{v_i}]) \models \phi(v) \wedge \psi(v)\})$$
$$\geq p \times \mu^k(\{(m_1, \ldots, m_k) \in M^k \mid$$
$$((\mathcal{M}, I), v[\overset{k}{m} / \overset{k}{v_i}]) \models \psi(v)\}).$$

Similarly, it will be required that for a probabilistic structure for $a$

$$(\mathcal{M}, I, \nu_a) \models \text{prob}(\phi[a] \mid \psi[a]) \geq p$$

only holds, if

$$\nu_a(\{(m_1, \ldots, m_n) \in M^n \mid (\mathcal{M}, I) \models \phi[m] \wedge \psi[m]\})$$
$$\geq \; p \times \nu_a(\{(m_1, \ldots, m_n) \in M^n \mid (\mathcal{M}, I) \models \psi[m]\}).$$

However, this mere *satisfaction of the constraints* in $\Phi^{\beta(a)}$ is insufficient for $\nu_a$, as it does not establish any connection between the measures $\mu^n$ and $\nu_a$. If the intuition is to be formalized, that $\nu_a$ should resemble $\mu^n$ as much as possible within the limits drawn by $\Phi^{\beta(a)}$, then something more is required.

First of all, the notion of "resemblance" has to be made precise. To this problem the following section is dedicated.

### 3.2 CROSS ENTROPY

Cross entropy ([Kul59]) commonly is interpreted as a "measure of information dissimilarity" for two probability measures [Sho86]. Usually, cross entropy is used in a rule to update a prior estimate for the probability distribution of some variable to a posterior estimate when some new information about the variable's actual distribution has been obtained. However, both its information theoretic interpretation and its unique properties make cross entropy also the most promising tool for bridging the gap between general statistical knowledge and subjective beliefs.

For a $\sigma$-algebra $\mathfrak{M}$ the set of probability measures on $\mathfrak{M}$ is denoted by $\Delta\mathfrak{M}$. Let $\mu, \nu \in \Delta\mathfrak{M}$ with $\nu \ll \mu$, i.e. for every $A \in \mathfrak{M}$: $\mu(A) = 0 \Rightarrow \nu(A) = 0$. In this case there exists a density function $f$ for $\nu$ with respect to $\mu$, and the cross entropy of $\nu$ with respect to $\mu$ can be defined by

$$CE(\nu, \mu) = \int f \ln f \, d\mu.$$

If $\nu \not\ll \mu$, define $CE(\nu, \mu) := \infty$. For $\mu \in \Delta\mathfrak{M}$ and a closed (with regard to the variation distance [1]) and convex subset $N \subseteq \Delta\mathfrak{M}$, which contains at least one $\nu$ with $CE(\nu, \mu) < \infty$, there is a unique $\nu_0 \in N$ such that $CE(\nu_0, \mu) < CE(\nu, \mu)$ for all $\nu \in N$, $\nu \neq \nu_0$ [Csi75]. Denote this $\nu_0$ by $\pi_N(\mu)$.

In the case that N is defined by a finite set of constraints

$$\{\nu(A_i) \geq p \times \nu(B_i) \mid$$
$$A_i, B_i \in \mathfrak{M}, \; p_i \in [0, 1], \; i = 1, \ldots, k\}$$

the following theorem reduces the problem of computing $\pi_N(\mu)$ to a CE-minimization on a finite probability space.

For a subalgebra $\mathfrak{M}' \subseteq \mathfrak{M}$ and $\mu \in \Delta\mathfrak{M}$ the notation $\mu \restriction \mathfrak{M}'$ is used for the restriction of $\mu$ to $\mathfrak{M}'$. $N \restriction \mathfrak{M}'$ stands for $\{\nu \restriction \mathfrak{M}' \mid \nu \in N\}$.

**Theorem 3.3** Let $\mathfrak{M}^0$ be a finite subalgebra of $\mathfrak{M}$ generated by a partition $\{A_1, \ldots, A_k\} \subseteq \mathfrak{M}$ of M. Let $\mu \in \Delta\mathfrak{M}$, and $N \subseteq \Delta\mathfrak{M}$ be defined by a set of constraints on $\mathfrak{M}^0$, i.e. for all $\nu \in \Delta\mathfrak{M}$:

$$\nu \in N \; \Leftrightarrow \; \nu \restriction \mathfrak{M}^0 \in N \restriction \mathfrak{M}^0.$$

Then $\pi_N(\mu)$ is defined iff $\pi_{N \restriction \mathfrak{M}^0}(\mu \restriction \mathfrak{M}^0)$ is defined, and in this case for every $C \in \mathfrak{M}$:

$$\pi_N(\mu)(C) = \sum_{i=1}^{k} \pi_{N \restriction \mathfrak{M}^0}(\mu \restriction \mathfrak{M}^0)(A_i) \, \mu(C \mid A_i).$$

The proof of this theorem is basically an application of property 9 from [SJ81].

**Corollary 3.4** Let $\{A_1, \ldots, A_k\} \subseteq \mathfrak{M}$ be a partition of M, let N be defined by a set of constraints

$$\{\nu(A_i) = p_i \mid i = 1, \ldots, k\}, \quad \sum_{i=1}^{k} p_i = 1.$$

Then, for every probability measure $\mu$ on $\mathfrak{M}$ with $p_i > 0 \Rightarrow \mu(A_i) > 0$ $(i = 1, \ldots, k)$, $\pi_N(\mu)$ is the probability measure obtained by applying Jeffrey's rule to $\mu$ and the given constraints.

Corollary 3.4 is a first indication that cross entropy might be the appropriate tool to model default reasoning about probabilities, and may serve as a preliminary justification for making cross entropy minimization the central element of the semantics for $L_S^g \cup L_S^{\beta(a)}$ now to be defined.

### 3.3 THE FINAL SEMANTICS

**Definition 3.5** Let $(\mathcal{M}, I, \nu_a)$ be a probabilistic S-structure for $a$, $KB = \Phi^\sigma \cup \Phi^{\beta(a)}$ a knowledge base. $(\mathcal{M}, I, \nu_a)$ is a *model* of KB iff

- $(\mathcal{M}, I) \models \Phi^\sigma$ as defined in section 3.1.

- With $Bel(a)$ the set of probability measures on $\mathfrak{M}^n$ that satisfy the constraints in $\Phi^{\beta(a)}$:

$$\nu_a = \pi_{Bel(a)}(\mu^n).$$

---

[1] The variation distance of $\nu_1, \nu_2 \in \Delta\mathfrak{M}$ is defined as the integral $\int |f_1 - f_2| \, d\nu$ where $\nu \in \Delta\mathfrak{M}$ is such that $\nu_1 \ll \nu$ and $\nu_2 \ll \nu$ (e.g. $\nu = 1/2(\nu_1 + \nu_2)$), and $f_i$ is a density for $\nu_i$ with respect to $\nu$.

$Bel(a)$ always is a closed (in the topology defined by the variation distance) and convex subset of $\Delta \mathfrak{M}^n$. To make sure that this will be the case is the reason for the restrictive syntax of $L_S^{\beta(a)}$. If it was allowed to express $\neg \text{prob}(\phi \mid \psi) \geq p$ for instance, then $Bel(a)$ would need no longer be closed. Permitting disjunctions $\text{prob}(\ldots) \geq p \vee \text{prob}(\ldots) \geq q$ destroys convexity. Hence, by the remarks in section 3.2, there exists a measure $\nu_a$ satisfying the condition of definition 3.5 iff $Bel(a)$ contains at least one measure $\nu$ with $\nu \ll \mu^n$. When this is not the case, then the statistical S-structure $(\mathcal{M}, I)$ can not be extended to a model of KB. Should this be the case for all $(\mathcal{M}, I) \models \Phi^\sigma$, then KB does not have a model.

Note that $Bel(a)$ is defined by constraints on the finite subalgebra of $\mathfrak{M}^n$ generated by the finitely many subsets of $M^n$ defined by the formulas appearing in $\Phi^{\beta(a)}$. Hence, theorem 3.3 applies to $\pi_{Bel(a)}(\mu^n)$, and even though $\mu$ and $\nu_a$ generally are probability measures on infinite probability spaces, cross entropy minimization only has to be performed on finite probability spaces.

The logic defined by definitions 2.1, 2.2 and 3.5 is denoted $\mathscr{L}^{\sigma\beta}$.

For a knowledge base KB and a sentence $\theta \in L_S^\sigma \cup L_S^{\beta(a)}$ the relation $\text{KB} \models \theta$ is defined as usual.

$\mathscr{L}^{\sigma\beta}$ is monotonic with respect to $\Phi^\sigma$, but non-monotonic with respect to $\Phi^{\beta(a)}$: if $\tilde\Phi^\sigma \supseteq \Phi^\sigma$ and $\Phi^\sigma \cup \Phi^{\beta(a)} \models \theta$, then $\tilde\Phi^\sigma \cup \Phi^{\beta(a)} \models \theta$ for every $\theta \in L_S^\sigma \cup L_S^{\beta(a)}$. If, on the other hand, $\tilde\Phi^{\beta(a)} \supseteq \Phi^{\beta(a)}$, then $\Phi^\sigma \cup \Phi^{\beta(a)} \models \theta$ does not imply $\Phi^\sigma \cup \tilde\Phi^{\beta(a)} \models \theta$.

## 4   WHY CROSS ENTROPY?

Cross entropy minimization, in the past, has received a considerable amount of attention as a rule for updating probability measure. Notably, Shore and Johnson have provided an axiomatic description of minimum cross entropy updating [SJ80], [SJ83]. They show that, if a function $f$ is used to define for a closed and convex set N of continuous or discrete probability measures and a prior $\mu$:

$$\pi_N^f(\mu) := \{\nu \in N \mid f(\nu, \mu) = \inf\{f(\nu', \mu) \mid \nu' \in N\}\},$$

and the mapping $(\mu, N) \mapsto \pi_N^f(\mu)$ satisfies a set of five axioms, then the function $f$ must in fact be equivalent to cross entropy.

It is beyond the scope of this paper to also give an axiomatic justification for putting cross entropy minimization at the core of definition 3.5 by formulating a set of conditions that the consequence relation $\models$ for $\mathscr{L}^{\sigma\beta}$ should satisfy, and then show that only cross entropy minimization will fulfill these conditions. Instead, the two theorems contained in this section demonstrate that using cross entropy leads to very desirable properties for $\mathscr{L}^{\sigma\beta}$, and indicate, when looked at as axioms rather than theorems, what an axiomatic justification for the use of cross entropy in the semantics of $\mathscr{L}^{\sigma\beta}$ would look like.

The two theorems are directly derived from the two central axioms in [SJ80], *subset independence* and *system independence*. The first one rephrases the property of subset independence to a statement about logical entailment in $\mathscr{L}^{\sigma\beta}$.

**Theorem 4.1** Let $\phi_1(v), \ldots, \phi_k(v) \in L_S$. Let KB= $\Phi^\sigma \cup \Phi^{\beta(a)}$ with

$$\Phi^\sigma \models \forall v(\phi_1(v) \dot\vee \ldots \dot\vee \phi_k(v))$$

(here $\dot\vee$ is the exclusive disjunction). Let

$$\begin{aligned}\Phi^{\beta(a)} &= \{\text{prob}(\phi_i[a]) \geq p_i \mid i = 1, \ldots, k\} \\ &\quad \cup \Phi_1^{\beta(a)} \cup \ldots \cup \Phi_k^{\beta(a)},\end{aligned}$$

where each $\Phi_i^{\beta(a)}$ is of the form

$$\{\text{prob}(\psi_{ij}[a] \mid \phi_{ij}[a]) \geq p_{ij} \mid j = 1, \ldots, l_k\}$$

for some $\phi_{ij}$ with $\Phi^\sigma \models \phi_{ij} \to \phi_i$. Then, for every $i \in \{1, \ldots, k\}$ and every subjective probability formula $\theta$ of the form $\text{prob}(\psi[a] \mid \phi_i[a]) \geq p$:

$$\Phi^\sigma \cup \Phi_i^{\beta(a)} \models \theta \quad \Rightarrow \quad \text{KB} \models \theta.$$

By theorem 4.1, reasoning by cases is possible in $\mathscr{L}^{\sigma\beta}$ under certain circumstances: if $\Phi^{\beta(a)}$ contains subjective beliefs that are each conditioned on one of several mutually exclusive hypotheses for $a$, then valid inferences about subjective beliefs conditioned on one of these hypotheses can be made by ignoring the information about the other hypotheses.

**Example 4.2** The prospects for a happy ending of the mystery film we have been watching not being very bright, we switch to a different channel where another film is running. This one can be easily identified as an American production, but it could be either a romance or a mystery:

$$\text{prob}(\text{American}f2 \wedge \text{Romance}f2) = 0.5, \quad (8)$$
$$\text{prob}(\text{American}f2 \wedge \text{Mystery}f2) = 0.5. \quad (9)$$

Also, we are ready to believe that

$$\text{prob}(\text{Happy\_end}f2 \mid \text{American}f2 \wedge \text{Romance}f2) = 0.95 \quad (10)$$

Suppose we are interested in estimating

$$\text{prob}(\text{Happy\_end}f2 \mid \text{American}f2 \wedge \text{Mystery}f2). \quad (*)$$

Before we are able to apply the statistical rule (2) in order to obtain this estimate, we make the additional observation that should $f2$ be a mystery, then it is not particularly likely to contain action scenes, contrary to what we generally expect from mystery films:

$$\text{prob}(\text{Action}f2 \mid \text{American}f2 \wedge \text{Mystery}f2) = 0.5, \quad (11)$$
$$[\text{Action}v \mid \text{American}v \wedge \text{Mystery}v]_{(v)} = 0.7. \quad (12)$$

This information is relevant for our estimate of (∗) because the existence of action scenes is correlated to a happy end by

[Action$v$ |

Americanv ∧ Mysteryv ∧ ¬Happy_end$v$]$_{(v)}$ = 0.5.    (13)

Let $KB_{f2} = \tilde{\Phi}^\sigma_{Movies} \cup \Phi^{\beta(f2)}$ consist of $\Phi^\sigma_{Movies}$ from example 2.3 and the new sentences (8)-(13). $KB_{f2}$ is of the form defined in theorem 4.1 with

$$\phi_1(v) = \text{American}v \wedge \text{Romance}v,$$
$$\phi_2(v) = \text{American}v \wedge \text{Mystery}v, \text{ and}$$
$$\phi_3(v) = \neg(\phi_1(v) \vee \phi_2(v)).$$

By theorem 4.1 we know that everything we can infer about (∗) from the smaller knowledge base obtained by removing (8)-(10) from $KB_{f2}$, also is valid with respect to $KB_{f2}$. By elementary computations it can be seen that

$\tilde{\Phi}^\sigma_{Movies} \models$ [Happy_end$v$ |

Americanv ∧ Mysteryv ∧ Action$v$]$_{(v)} = \frac{6}{7}$,    (14)

$\tilde{\Phi}^\sigma_{Movies} \models$ [Happy_end$v$ |

Americanv ∧ Mysteryv ∧ ¬Action$v$]$_{(v)} = \frac{2}{3}$.    (15)

Hence, with theorem 3.3

$KB_{f2} \models$ prob(Happy_end$f2$ | American$f2$ ∧ Mystery$f2$)
$$= 0.5 \times \frac{6}{7} + 0.5 \times \frac{2}{3} = \frac{16}{21}.    (16)$$

Also, combining (8)-(10) and (16) we get

prob(Happy_end$f2$)
$$= 0.5 \times 0.95 + 0.5 \times \frac{16}{21} \approx 0.856.    (17)$$

The following theorem is derived from the *system independence* axiom.

**Theorem 4.3** Let $KB = \Phi^\sigma \cup \Phi^{\beta(a)}$, where

$$\Phi^{\beta(a)} = \Phi^{\beta(a_0,\ldots,a_{k-1})} \cup \Phi^{\beta(a_k,\ldots,a_{n-1})},$$

i.e. the set of subjective probability formulas for $a$ consists of two disjoint sets for $(a_0,\ldots,a_{k-1})$ and $(a_k,\ldots,a_{n-1})$. Suppose that

$\Phi^\sigma \cup \Phi^{\beta(a_0,\ldots,a_{k-1})} \models$

prob($\phi_1[a_0,\ldots,a_{k-1}] \mid \psi_1[a_0,\ldots,a_{k-1}]$) ≥ $p_1$,    (18)

$\Phi^\sigma \cup \Phi^{\beta(a_k,\ldots,a_{n-1})} \models$

prob($\phi_2[a_k,\ldots,a_{n-1}] \mid \psi_2[a_k,\ldots,a_{n-1}]$) ≥ $p_2$.    (19)

Then

$KB \models$ prob($\phi_1[a_0,\ldots,a_{k-1}] \wedge \phi_2[a_k,\ldots,a_{n-1}]$

$\mid \psi_1[a_0,\ldots,a_{k-1}] \wedge \psi_2[a_k,\ldots,a_{n-1}]$) ≥ $p_1p_2$.    (20)

Theorem 4.3 remains true, when the inequality in (18)-(20) is replaced with equality.

**Corollary 4.4** For KB as in the preceding theorem and for every subjective belief formula $\theta \in L_S^{\beta(\tilde{a}^k)}$:

$$\Phi^\sigma \cup \Phi^{\beta(a_0,\ldots,a_{k-1})} \models \theta \;\Rightarrow\; KB \models \theta.$$

Roughly speaking, theorem 4.3 states, that when $\Phi^{\beta(a)}$ does not contain any information connecting one constant $a_i$ with another constant $a_j$, then these constants are interpreted as independent. Especially, ignoring the information about $a_j$ still leads to valid inferences about $a_i$.

**Example 4.5** Ultimately, we want to know which of the two films $f1$ and $f2$ is likely to be the better one. Better is a predicate for which we have the axioms

$$\forall v_0 \neg \text{Better}v_0 v_0 \qquad (21)$$
$$\forall v_0 v_1 (v_0 \neq v_1 \rightarrow (\text{Better}v_0 v_1 \leftrightarrow \neg \text{Better}v_1 v_0)) \qquad (22)$$

and a useful statistic:

[Better$v_0 v_1$ | Happy_end$v_0$ ∧ ¬Happy_end$v_1$]$_{(v_0,v_1)}$
$$= 0.95.    (23)$$

Let $KB_{f1f2}$ be the union of $KB_{f1}$, $KB_{f2}$ and the sentences (21)-(23). From (21)-(23)

[Better$v_0 v_1$ | $v_0 \neq v_1$ ∧ Happy_end$v_0$

∧ Happy_end$v_1$]$_{(v_0,v_1)}$ = 0.5    (24)

[Better$v_0 v_1$ | $v_0 \neq v_1$ ∧ ¬Happy_end$v_0$

∧ ¬Happy_end$v_1$]$_{(v_0,v_1)}$ = 0.5    (25)

[Better$v_0 v_1$ | ¬Happy_end$v_0$

∧ Happy_end$v_1$]$_{(v_0,v_1)}$ = 0.05    (26)

can be derived by exploiting the fact that we are dealing with product measures, and therefore, for all $p \in [0,1]$:

$\models$ [Better$v_0 v_1$ | $v_0 \neq v_1$ ∧ Happy_end$v_0$

∧ Happy_end$v_1$]$_{(v_0,v_1)} \geq p$

↔ [Better$v_1 v_0$ | $v_0 \neq v_1$ ∧ Happy_end$v_0$

∧ Happy_end$v_1$]$_{(v_0,v_1)} \geq p$.

By our previous results (1) (formally justified by corollary 3.4) and (17), and theorem 4.3, the probabilities of the conditioning events in (23)-(26) for $f1$ and $f2$ are known to be $0.32 \times (1 - 0.856) = 0.046$, $0.32 \times 0.856 = 0.274$, $(1 - 0.32) \times (1 - 0.856) = 0.098$ and $(1 - 0.32) \times 0.856 = 0.582$ respectively. One final application of Jeffrey's rule, sanctioned by theorem 3.3, then yields

$KB_{f1f2} \models$ prob(Better$f1f2$) =
$$0.95 \times 0.046 + 0.5 \times 0.274 +$$
$$0.5 \times 0.098 + 0.05 \times 0.582 = 0.259,$$

which is a suitable result to settle the question about which film we are going to watch.

Obviously, this example has been an extremely simple illustration of the given definitions and theorems throughout: neither will it be possible, in more realistic examples, to reduce cross entropy minimization to an application of Jeffrey's rule, nor will the resulting probabilities usually be unique values rather than intervals.

## 5 RELATED WORK

In [PV89] and [PV92] Paris and Vencovská consider basically the same inference problem as is discussed in the present paper. They assume that two types of probabilistic constraints on expressions in propositional logic are given: one type referring to general proportions, the other to subjective beliefs about an individual. Their approach to dealing with the dichotomy of the probabilistic information is quite different from the one here presented: it is proposed to transform the constraints on the subjective beliefs about an object $a$ to statistical constraints conditioned on a newly introduced propositional variable A representing an ideal reference class for $a$, i.e. the set of all elements that are "similar to" $a$. Then an additional constraint is added that the absolute probability of this set is very small. Thus, all the constraints can be viewed as being on one single probability distribution. Paris and Vencovská then explore different inference processes that can be applied to these constraints in order to obtain a single probability distribution on the propositional formulas. Most notably, they consider the maximum entropy approach, and show that when it is used the resulting conditional probability distribution on the variable A is just the distribution on the formulas not containing A that minimizes cross entropy with respect to the global distribution on these formulas under the constraints for $a$ (more precisely, this will be the case for the limiting distribution when the absolute probability of A tends to zero).

The techniques of probabilistic inference explored by Paris and Vencovská are quite different from the one discussed in this paper in that, as demanded by the uniform encoding of statistical and subjective probabilities, one process of inference is applied to both kinds of information simultaneously. This makes Paris and Vencovská's paradigm for probabilistic inference a somewhat less likely framework for default reasoning about probabilities, where it is the key issue to give an interpretation of the subjective beliefs as a function of the interpretation of the statistical information.

However, the mere semantic principle of interpreting subjective beliefs via conditional probabilities on a new reference class also allows for a separate processing of the constraints given for the domain in general and the constraints given with respect to the reference class. Thus, the two approaches of interpreting the subjective beliefs held about an object as either the conditional distribution on a special reference class, or as an alternative measure on the domain as in $\mathscr{L}^{\sigma\beta}$, basically allow for the same scope of probabilistic reasoning. If it is intended, though, to clearly distinguish the reasoning about the statistics from the reasoning about beliefs — a separation pushed to the extreme in the probabilistic logics of Bacchus et al. and Halpern — the second approach probably will lead to greater conceptual clarity.

## 6 CONCLUSION

$\mathscr{L}^{\sigma\beta}$ is a logic that models the forming of subjective beliefs about objects on the basis of statistical information about the domain and already existing beliefs. The novelty of the approach here presented lies in the idea of interpreting constant symbols as probability measures over the domain, which leads to semantics that seem to be better suited to describe the interaction of statistical and belief probabilities than possible worlds semantics. In order to make effective use of cross entropy minimization, a fairly restrictive syntax with regard to expressing subjective beliefs was introduced.

It should be pointed out, though, that $\mathscr{L}^{\sigma\beta}$ is open to generalizations in various ways. Disjunctions and negations of subjective probability sentences might be allowed, in which case the condition $\nu_a = \pi_{Bel(a)}(\mu^n)$ in definition 3.5 has to be replaced by the demand that $\nu_a$ is one of the measures in the closure of $Bel(a)$ that minimizes cross entropy with respect to $\mu$.

Also, interpreting constant symbols as probability measures over the domain is a feasible way to interpret formulas in which statements of subjective belief and statistical relations are arbitrarily nested, thus allowing to express statements like

$$[\text{prob}(\text{Better} f1 v) \geq 0.9]_{(v)} \geq 0.2$$

("for some ($\geq 0.2$) $v$ it is believed that $f1$ is very likely ($\geq 0.9$) to be better than $v$). When formulas like these are allowed, however, it is more difficult to define what their proper default interpretation should be, because the interaction of statistical information and subjective beliefs can no longer be viewed as essentially one-way only.

### Acknowledgement

The author is greatful for some helpful remarks and suggestions received from an anonymous referee. Particularly, they contained a valuable clarification regarding the interrelation of direct inference and Jeffrey's rule.